\documentclass{article}

\usepackage{microtype}
\usepackage{graphicx}
\usepackage{subfigure}
\usepackage{booktabs} 

\usepackage{hyperref}

\usepackage{amsmath}
\usepackage{amssymb}
\usepackage{mathtools}
\usepackage{amsthm}

\usepackage[capitalize,noabbrev]{cleveref}

\usepackage{natbib}

\usepackage{rotating}
\usepackage{amsmath}
\usepackage{mathptmx}
\usepackage{wrapfig}
\RequirePackage{algorithm}
\RequirePackage{algorithmic}
\usepackage{tabularx}
\usepackage{booktabs}
\usepackage{multicol}

\newcommand{\citewithauthor}[1]{\citeauthor{#1}~\cite{#1}}

\usepackage{graphicx}

\newcommand{\bA}{\mathbf{A}}

\newcommand{\bX}{\mathbf{X}}

\newcommand{\bXi}{\boldsymbol{\Xi}}
\newcommand{\bxi}{\boldsymbol{\xi}}

\newcommand{\bTheta}{\boldsymbol{\Theta}}

\theoremstyle{plain}

\theoremstyle{definition}

\theoremstyle{remark}

\usepackage[textsize=tiny]{todonotes}

     \usepackage[preprint]{neurips_2021}

\newcommand{\affnum}[1]{{\normalsize\rm\textsuperscript{\,#1}}}
\newcommand{\affiliation}[2]{\normalsize\rm\textsuperscript{#1}#2}

\author{%
    { Rushiv Arora }\affnum{1} 
    \And
    { Bruno Castro da Silva }\affnum{1} \And
    { Eliot Moss }\affnum{1} \thanks{Correspondence to: Rushiv Arora \texttt{<rrarora@umass.edu>}, Eliot Moss \texttt{<moss@cs.umass.edu>}} \And
    \affiliation{1}{Manning College of Information and Computer Sciences \\
     University of Massachusetts Amherst \\ Amherst MA 01003, USA
    } \\
}

\begin{document}
\title{Model-Based Reinforcement Learning with SINDy}
\maketitle





\begin{abstract}
We draw on the latest advancements in the physics community to propose a novel
method for discovering the governing non-linear dynamics of physical systems
in reinforcement learning (RL).  We establish that this method is capable of
discovering the underlying dynamics using significantly fewer trajectories (as
little as one rollout with $\leq 30$ time steps) than state of the art model
learning algorithms.  Further, the technique learns a model that is accurate
enough to induce near-optimal policies given significantly fewer trajectories
than those required by model-free algorithms.  It brings the benefits of
model-based RL without requiring a model to be developed in advance, for
systems that have physics-based dynamics.

To establish the validity and applicability of this algorithm, we conduct
experiments on four classic control tasks.  We found that an optimal policy
trained on the discovered dynamics of the underlying system can generalize
well.  Further, the learned policy performs well when deployed on the actual
physical system, thus bridging the model to real system gap.  We further
compare our method to state-of-the-art model-based and model-free approaches,
and show that our method requires fewer trajectories sampled on the true
physical system compared other methods.  Additionally, we explored approximate
dynamics models and found that they also can perform well.


\end{abstract}

\section{Introduction}

In reinforcement learning, it is generally held that model-based approaches
learn more quickly than do model-free approaches assuming the model is
accurate enough \cite{Kaelbling+:1996:ArXiv}.  However, it is well known that
learning model is challenging in general.  We show here that for systems with
underlying physics dynamics, such as robotic control, one can often learn an
accurate low-dimension model quickly, using very little training data from the
actual system.  Training with that model can result in asymptotic performance
comparable to model-free approaches, using models with many fewer parameters
and that are trained using less data than state of the art model-based
algorithms.  Thus, for systems with physics-like dynamics, we bridge the gap
between model-free and model-based RL.

\newcommand{\vect}[1]{\overset{\rightharpoonup} #1}
\section{SINDy}

SINDy, which stands for Sparse Identification of Non-linear Dynamics, is an
approach developed in the physics community for extracting equational models
of physical systems from time series data.  Specifically, SINDy can extract
differential equations (ODEs, PDEs) or difference equations from data given a
collection of possible equation terms (generally called \emph{features} in the
ML community) that are functions of the input values.  Such terms might
include polynomial and trigonometric functions of the data values.  The
``Sparse'' of SINDy indicates that it tries to extract from a possibly large
space of terms the minimum number of ones necessary for an accurate model.
``Non-linear'' indicates that the terms (features) may be non-linear functions
of one or more input values.  SINDy was introduced by
\citewithauthor{Brunton+:2016:PNAS}, who showed it is powerful enough to
extract the physics even of chaotic systems such as the True Lorentz System
\cite{Brunton+:2016:PNAS}.  Similar forms apply to discrete-time and noisy
systems \cite{Brunton+:2016:PNAS}.  \citet{Brunton+:2016:IFAC} extends that
work by extended SINDy to deal with force-driven systems (control), and
\cite{Boninsegna+:2018:JCP} developed a stochastic version.  To demonstrate
the power of the force-driven extension they solved the Lotka-Volterra
predator-prey model and the Lorenz system with forcing and control.

SINDy extracts a dynamics model with applied actions by solving the equation
\begin{equation}
\dot{x}(t) = f(x(t);a(t))
\end{equation}
for $f$, where the vector
\begin{equation}
x(t) = [x_1(t), x_2(t), \dots, x_n(t)]^T \in R^n
\end{equation}
represents the observation of the system at time $t$, the vector
\begin{equation}
a(t) = [ a_1(t), a_2(t), \dots, a_k(t)]^T \in R^k
\end{equation}
the action (typically physical forces) applied to the system at time $t$, the
semicolon indicates vector concatenation, and the (possibly nonlinear)
function $f(x(t);a(t))$ represents the dynamic constraints that define the
equations of motion of the system.  We write $f_i$ for the function that
defines $\dot{x_i}(t)$.  SINDy uses a similar form for discrete-time
difference equations.  SINDy can be extended to probabilistic models, but that
was not necessary in our application.


At the heart of SINDy lies a method for feature selection and sparse
regression, based on the principle that only a few terms in the regression
model will be important.  By using intuition about the model, the user
proposes a collection of feature functions, which may include polynomials,
Fourier terms, etc., that the user \emph{thinks} might govern the dynamics.
Each feature is a possibly non-linear function of one of more of the input
variables, that is, of the $x_i$ and $a_i$.  SINDy attempts to extract a model
where each $f_i$ is a \emph{linear} function of the features.

To that end, SINDy defines
\begin{equation}
\hat{\bTheta} = [\bTheta_1, \bTheta_2, \dots, \bTheta_F], \mbox{where}\,\,\bTheta_i \in
R^{n+k} \rightarrow R,
\end{equation}
a vector of functions of the $x_i$ and $a_i$ that we will call \emph{feature
functions}.  While the $\bTheta_i$ take $n+k$ arguments, they typically depend
on only a few elements of $x(t);a(t)$.  $F$ is the number of features.  We
further define
\begin{equation}
\bTheta(x;a) = [\bTheta_1(x;a), \bTheta_2(x;a), \dots, \bTheta_F(x;a)].
\end{equation}

SINDY also defines an $n \times F$ matrix $\bXi$ of real values $\bxi_{i.j}$
in order to define
\begin{equation}
f_i(x;a) = \bTheta(x;a) \cdot \bXi_i.
\end{equation}
Thus the $f_i$ are indeed linear functions of the possibly non-linear
features, with $\bXi$ giving the coefficients (weights) of those features for
each $f_i$.  We can now write
\begin{equation}
f(x;a) = \bTheta(x;a) \cdot \bXi
\end{equation}
for the overall definition of $f$.

We now proceed to define the optimization problem that SINDy solves.  Given a
sequence of observations $x(1), x(2), \dots, x(N)$ and actions $a(1), a(2),
\dots, a(N)$, SINDy can compute actual derivatives $\dot{x}(t)$ according to
several methods, and can also take user-supplied values for those derivatives
or a user-supplied derivative calculating function.  We used its built-in
smoothed finite differencing method.  We can thus form pairs for training
$(\dot{x}(i), x(i);a(i))$.  Note that the $x(t);a(t)$ inputs are readily
extracted from RL trajectories of the typical $(s,a,s')$ form.

We aggregate the $x$ and $a$ values into arrays $X$ and $A$, respectively:
\begin{eqnarray}
\bX &= \overset{\text{\normalsize state}}{\left.\overrightarrow{\begin{bmatrix}
x_1(1) & x_2(1) & \cdots & x_n(1)\\
x_1(2) & x_2(2) & \cdots & x_n(2)\\
\vdots & \vdots & \ddots & \vdots \\
x_1(N) & x_2(N) & \cdots & x_n(N)
\end{bmatrix}}\right\downarrow}\begin{rotate}{270}\hspace{-.125in}time~~\end{rotate}\label{Eq:DataMatrix},
\end{eqnarray}
\begin{eqnarray}
\bA &= \overset{\text{\normalsize action}}{\left.\overrightarrow{\begin{bmatrix}
a_1(1) & a_2(1) & \cdots & a_k(1)\\
a_1(2) & a_2(2) & \cdots & a_k(2)\\
\vdots & \vdots & \ddots & \vdots \\
a_1(N) & a_2(N) & \cdots & a_k(N)
\end{bmatrix}}\right\downarrow}\begin{rotate}{270}\hspace{-.125in}time~~\end{rotate}\label{Eq:DataMatrixA},
\end{eqnarray}
and we write $X;A$ for the $(n+k) \times N$ array formed by appending each
$a(i)$ to the corresponding $x(i)$.  We extend our $\bTheta$ notation to
define
\begin{equation}
\bTheta(X;A) = [\bTheta(x(1);a(1)), \bTheta(x(2);a(2)), \dots, \bTheta(x(N);a(N))].
\end{equation}
The optimization problem to be solved is then:
\begin{equation}
\dot{X} = \bTheta(X;A) \cdot \bXi,
\end{equation}
and we desire a solution where $\bXi$ is sparse.  Note that this is now a
sparse \emph{linear} regression problem, in terms of the (possibly non-linear)
functions $\bTheta$ of the input data $X;A$.  SINDy can apply any of a variety
of sparse regression methods.  We use its Sequentially Thresholded Least
Squares (STLSQ) method, which uses Ridge regression, with a threshold of
$0.0009$.  It iteratively solves the least squares regression problem with a
regularizer being the L2 norm of the weights $\bXi$, masking out weights below
the threshold (setting them to $0$).

Let us consider the very small example of a mass $M$ moving in one dimension
$x$ under a time varying action force $g$.  (We use $g$ to avoid confusion
with $f$.)  Our observation are the position $x$ and velocity $v$.  From
Newton's Law we know that $\dot{v} = g/M$, so the equations of motion are:

\begin{equation}
\begin{bmatrix}
\dot{x} \\
\dot{v}
\end{bmatrix} =
\begin{bmatrix}
0 & 1 & 0 \\
0 & 0 & 1/M \\
\end{bmatrix} \cdot
\begin{bmatrix}
x \\
v \\
g
\end{bmatrix}
\end{equation}
In this case, given suitable data, and a collection of feature functions
$\hat{\bTheta}$ that included $\lambda (x,v,g). v$ and $\lambda (x,v,g). g$ (or
more loosely, terms $v$ and $g$), SINDy should arrive at a solution $\bXi$
whose non-zero elements are exactly the $1$ and $1/M$ in the equations of
motion, to within computational error.  Notice that SINDy in effect discovers
the mass $M$, that is, we knew the \emph{form} of the equations, but not
necesarily the exact values of the coefficients.  This is important with real
world robots, each one of which will exhibit slight variations from a desired
specification, etc.

Here we knew the exact form in advance.  If we were less certain, we might
include more functions in $\hat{\bTheta}$, such as terms of the form $1$, $x$,
$x^2$, $x\cdot v$, $\sin x$, etc., and SINDy would still arrive at the same
solution because of its accuracy and sparseness.

The key insights of SINDy are:
\begin{itemize}
\item deriving a model of physics-based dynamics using physically plausible
  (but possibly nonlinear) features of the observations and actions; and
\item assuming a model that has a simple equational form rather than trying to
  learn a model via ``brute force'' function approximation.
\end{itemize}
Together these insights allow learning of a highly accurate model with a
relatively small number of observations.  Further, the models have only a
small number of parameters (typically much less than the number of elements of
$\bXi$, which itself has orders of magnitude fewer weights than a typical
neural net model).  One expects SINDy to do well if the problem is physics
based, that is, the problem admits of solution as a relatively simple,
possibly nonlinear, differential (or difference) equation.  It is not a
general solution for extracting a model from an arbitrary data set.

\section{Dyna-Style Learning with SINDy}

\begin{figure}[htb]
    \vspace*{-12pt}
    \begin{algorithm}[H]
    \label{alg:dyna}
        \caption{Dyna-Style Model-Based RL with SINDy}
       \begin{algorithmic}
           \STATE Hyper-parameters: Integers $N_e$ and $N$
           \STATE Initialize $\mathcal{D}_{\mathrm{\emph{env}}}$ and $\mathcal{D}_{\mathrm{\emph{SINDy}}}$ as empty data sets
           \STATE Initialize policy $\pi$ and SINDy parameters $\bXi$
           \STATE ~~~~to random values
           \FOR{$N_e$ \emph{rollouts}}
              \STATE Collect data ($s_i$, $a$, $s_{i+1}$) on real environment with
              \STATE ~~~~random or pseudo-random policy
              \STATE$\mathcal{D}_{\mathrm{\emph{SINDy}}} \gets \mathcal{D}_{\mathrm{\emph{SINDy}}} \bigcup$ ($s_i$, $a$, $s_{i+1}$)
            \ENDFOR
            \STATE Train model $\bXi$ on $\mathcal{D}_{\mathrm{\emph{SINDy}}}$ using SINDy
            \WHILE{$\pi$ \emph{is not optimal}}
                \FOR{$N$ \emph{epochs}}
                    \STATE Collect rollout $roll_{\mathrm{\emph{sim}}}$ from model $\bXi$
                    \STATE Train $\pi$ on simulated $roll_{\mathrm{\emph{sim}}}$ using
                    \STATE ~~~~model-free algorithm and known reward function
                \ENDFOR
                \STATE Collect a single rollout $r_{\mathrm{\emph{real}}}$ from real environment
                \STATE Train $\pi$ on $r_{\mathrm{\emph{real}}}$ using an arbitrary model-free algorithm
                \STATE $\mathcal{D}_{\mathrm{\emph{SINDy}}} \gets \mathcal{D}_{\mathrm{\emph{SINDy}}} \bigcup r_{\mathrm{\emph{real}}}$
            \ENDWHILE
       \end{algorithmic}
    \end{algorithm}
    \vspace*{-6pt}
\end{figure}

We propose a Dyna-style learning algorithm with SINDy at its base.  In this
algorithm SINDy learns an accurate sparse model of the non-linear dynamics of
a physical RL system.  We define two hyper-parameters that can be chosen
depending on the complexity of the physical system we are trying to learn:
\begin{itemize}
\item $N_e$: The number of rollouts for which the algorithms collects data
  to train SINDy.  The algorithm uses a random or pseudo-random policy for these
  rollouts.  SINDy does not generally need much data, so $N_e$ is typically
  small, and in fact a value of 1 sufficed in our experiments.
\item $N$: The number of epochs to train using data generated by the
  SINDy-induced model for each epoch of training using data from the actual
  system, i.e., the actual system is used only one out of every $N+1$ epochs.
  If the SINDy model is accurate over a wide enough part of the state space,
  $N$ can be set to an arbitrarily large value, which worked for a number of
  our experiments.
\end{itemize}
Notice that we assume $\hat{\bTheta}$ has been chosen in advance and thus
speak of the model as being $\bXi$, which strictly speaking is the
coefficients of the model.  The Dyna-style algorithm can readily be extended
to retrain the SINDy model periodically if there were benefit to doing so, but
it was not necessary in our experiments.

\section{Experiments}

We conducted experiments using four environments with three levels of
difficulty in mind: discrete actions (Cart Pole), continuous actions (Mountain
Car and Pendulum Swing Up), and realistic Mujoco physics control problems with damping and
friction (Inverted Pendulum).  We use the Python package PySindy open sourced by
\citewithauthor{Kaptanoglu+:2021:PySINDy}.  We selected variants of SINDy from
among its continuous-time, discrete-time, and driven models appropriate to
each experiment.  We use a state of the art model-free algorithm,
Soft Actor-Critic (SAC), introduced by \citewithauthor{Haarnoja+:2018:arXiv}
and extended by \citewithauthor{Christodoulou:2019:arXiv}, as our basis for
training on both the real system and the simulated rollouts described in
Algorithm~\ref{alg:dyna}.  We also compare our results against Model-Based
Policy Optimization (MBPO), a state of the art model-based method introduced
by \citewithauthor{Janner+:2019:NIPS}.  MBPO and our model differ primarily in
that MBPO learns a model represented by a neural net while SINDy learns a
model represented by a differential (or difference) equation.
Our method outperforms both
other methods in all experiments, giving us a $4$--$100\times$, $40\times$,
and $200\times$ speedup against MBPO, and $15$--$375\times$, $60\times$,
$500\times$, and $5\times$ speedup against SAC for the Inverted Pendulum,
Pendulum Swing Up, Mountain Car, and Cart Pole problems, respectively.

\begin{figure*}[t]
    \centering
    \includegraphics[width=\textwidth]{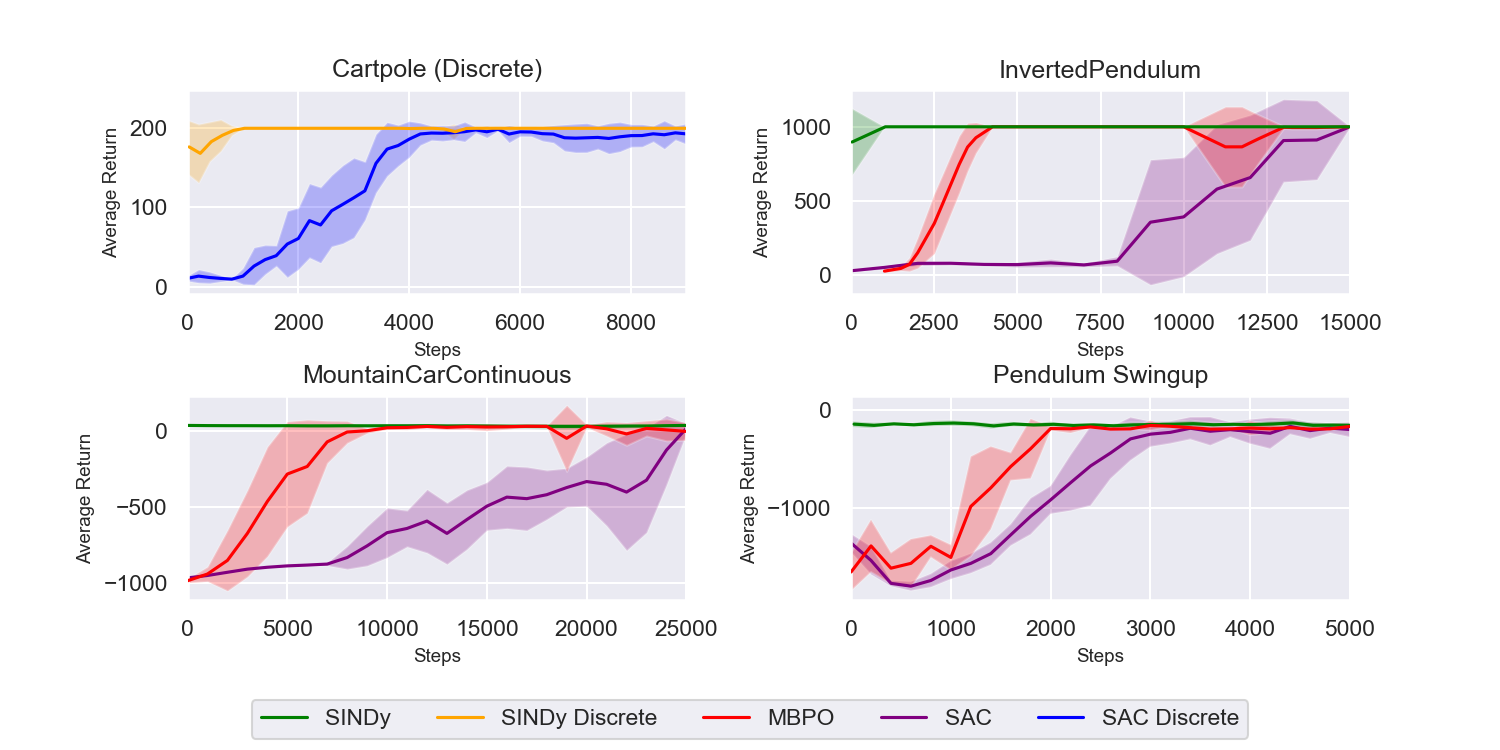}
    \caption{The results of our SINDy method compared to state of the art
      model-based (MBPO) and model-free (SAC) methods.  Our method outperforms
      current methods in discrete action, continuous action, and noisy/damped
      environments.  We found $4$--$100\times$, $40\times$, and $200\times$
      speedup against MBPO and $15$--$375\times$, $60\times$, $500\times$, and
      $5\times$ speedup against SAC for the Inverted Pendulum, Pendulum Swing
      Up, Mountain Car, and Cart Pole, respectively.}
    \label{fig:results}
    \vspace{-10pt}
\end{figure*}

Figure~\ref{fig:results} shows our experimental results.  Each experiment is
averaged over 10 different seeds, with further averaging performed by
evaluating the agent 10 times for each seed.  All results involve running the
true robotic system once for a small designated number of time steps
(Table~\ref{tab:hyperparameters}) to obtain samples for training the SINDy model, and then
performing further policy improvement steps based only on the
SINDy derived model. The plots for SINDy have been shifted right by the number of time steps
gathered to train SINDy based on interactions with the real environment.  We
now offer more details of each experiment.

\textbf{Discrete Classic:}  Our discrete action case is the Cart Pole
environment.  Data from a single rollout of 30 steps was sufficient for SINDy to
identify a high accuracy dynamics model.  To avoid rapid termination of an episode we
applied force in opposite directions at alternating time steps, with
occasional random actions taken for exploration.

We can summarize the dynamics of the Cart Pole system in these equations:

\begin{equation*}
  \ddot\theta = \frac{(g\cdot\sin{\theta} - \cos{\theta}\cdot\mbox{\emph{C}})}{l\cdot(\frac{4}{3} - \frac{m_p}{m_p + m_c}\cdot(\cos{\theta})^2)},
\end{equation*}
\begin{equation}
    \ddot x = \mbox{\emph{C}} - \frac{l}{m_p + m_c}\cdot\ddot\theta\cdot\cos{\theta}
\label{eq:cartpole}
\end{equation}
where
\begin{equation*}
  \mbox{\emph{C}} = (F + l\cdot\dot\theta^2\sin{\theta})/(m_p+m_c),
\end{equation*}

where $l$ is the length of the pole, $m_p$ its mass, $x$ the position of the
cart, $m_c$ its mass, $\theta$ the vertical angle between the two, and $F$
the force.  Using a small $\theta$ assumption, since the pole falls and the
episode terminates if the $\theta$ is not small, we can write $\sin{\theta}
\approx \theta$ and $\cos{\theta} \approx 1$, resulting in these equations:

\begin{equation*}
  \ddot\theta = \frac{(g\cdot\theta - \mbox{\emph{C}})}{l\cdot(\frac{4}{3} - \frac{m_p}{m_p + m_c})},
\end{equation*}
\begin{equation}
    \ddot x = \mbox{\emph{C}} - \frac{l}{m_p + m_c}\cdot\ddot\theta
\label{eq:cartpole_approx}
\end{equation}
where
\begin{equation*}
  \mbox{\emph{C}} = (F + l\cdot\dot{\theta}^2\cdot\theta)/(m_p+m_c),
\end{equation*}

Thus, substituting the right hand side of the equation for $\ddot{\theta}$
into the one for $\ddot{x}$, the right hand sides of the system of dynamics
equations can be written in terms of $\theta$, $\dot{\theta}$, $x$, $\dot{x}$,
and constants.  The $\hat{\bTheta}$ we provided to SINDy was $\begin{bmatrix}
  1, & a, & a^2, & a\cdot b, & a^2\cdot b\end{bmatrix}$ where $a$ and $b$ can be any of
$\theta$, $\dot{\theta}$, $x$, and $\dot{x}$.
Figure~\ref{fig:results} shows that the approximation is accurate enough that
quitting when we are optimal using the SINDy model is still accurate in the
real environment and needs just two further episodes of training in the real
environment for fine tuning.


\textbf{Continuous Classic:}  The equations governing the dynamics for the
continuous environments of
Mountain Car and Pendulum Swing Up are learned precisely by SINDy (to 4
decimal places), extracting appropriate features from among a larger set.  For
example, for Mountain Car, the features we used were $1$, $x$, $x^2$, and
$\sin kx$ and $\cos kx$ for $k=1,2,3$.  The results of
Figure~\ref{fig:results} show that the policy
that is optimal on the SINDy dynamics is also optimal in the real
environment.  In fact, when we examine the equations learned
by SINDy, we see that they \textbf{\emph{match the true dynamics}} of these
environments.  It is worth noting that the Mountain Car domain has inelastic
collisions, and our model is robust in the face of those discontinuities when learning
dynamics.  Furthermore SINDy learned these dynamics using a single rollout of
length 100 for Mountain Car and 20 for Pendulum Swing Up.

\textbf{Mujoco:}  We now consider the
Inverted Pendulum domain of the Mujoco physics simulator.  This adds the
challenge of having damping on the controller and friction added to the
system.  The dynamics that SINDy learned generalize well to the true environment as
can be seen in Figure~\ref{fig:results} figure, and it needed
\textbf{\emph{at most one additional training episode}} for fine tuning the policy
learned using the model so that it is optimal with respect to the true physical
system.  We used the same approximation here as for
Equation~\ref{eq:cartpole_approx} and used the same policy for collecting the 
initial samples.


\vspace{-5pt}
\begin{table*}[htb]  
\begin{center}
    \medskip
    \caption{Hyper-parameters used for training the SINDy model and Results}
    \label{tab:hyperparameters}
    \begin{tabular}{lcccccrr}\hline
                &       &  & &
    Dynamics &             & & \\
    Environment & $N_e$ & $R$  & $N$ &
    learned  & Generalizes & $P$ & $P'$ \\\hline\hline
    Cart Pole    & 1 & 30 & $\infty$ & Approximate & Yes & 164 &$\sim$70\\\hline
    Mountain Car & 1 & 50 & $\infty$ & Exact       & Yes &  50 &   7\\\hline
    Pendulum     &   &    &          &             &     &     &    \\
    Swing up     & 1 & 20 & $\infty$ & Exact       & Yes &  99 &  10\\\hline
    Inverted     &   &    &          &             &     &     &    \\
    Pendulum     & 1 & 30 & $\infty$ & Approximate & Yes & 164 &  50\\\hline
    \end{tabular}
\end{center}

\noindent
$N_e$ is the number of rollouts; $R$ is the length of each rollout; $N$ is the
number of episodes using just the model (vs.\ the actual system); $P$ is the
number of parameters (size of $\bXi$); $P'$ is the number of non-zero
parameters (Cart Pole is stochastic so the number varied a little).  The
MBPO model sized was: 613,036 parameters.
The SINDy model size was $n \cdot F$, where $n$ is the number of dimensions in the
state and $F$ the number of functions (features) in $\bXi$.
\end{table*}

\textbf{Discussion:} As Table~\ref{tab:hyperparameters} shows, our approach learns
models that have a very small number of parameters, particularly compared with
those learned by MBPO (613,036 parameters).  Furthermore, since our models
represent physics dynamics equations explicitly, they are highly
interpretable, while function approximation neural nets generally are not.  As
previously discussed, the models are very accurate and can be learned with
only a small amount of training data.  The method works for a range of kinds
of dynamics and control, both continuous and discrete.  We have further seen
that approximate dynamics, such as replacing $sin(\theta)$ by $\theta$ when
angles tend to be small in the region of the state space that is of interest,
can lead to dynamics models accurate enough for training to result in optimal
behavior.

\section{Conclusions and Future Work}

We presented a method for learning the model of physics based RL systems.  It is
capable of learning either exact or high-accuracy approximate non-linear dynamics
from small numbers of samples.  We showed results from four environments
demonstrating how training with these models results in asymptotic performance as good
as that achieved by state of the art model-free methods, while converging
significantly more rapidly and requiring less training data.  Our
dynamics models are of low dimension, easy to extract, and highly
interpretable, advantages they have over state of the art model-based methods.

In summary, our contributions are:
\begin{enumerate}
    \item Our algorithm matches or exceeds the asymptotic performance of existing
      state of the art model-based and model-free learning methods on these
      tasks while
      requiring significantly fewer time steps of interaction with the real
      system.  In our experiments, we needed at most 50 time steps of
      interaction with the real system to identify high accuracy models that
      allow induction of near optimal policies.  We reduced the time steps of
      interaction with the real system
      necessary to convergence by $4\times$--$100\times$, $40\times$, and
      $200\times$ against MBPO, and $15\times$--$375\times$, $60\times$,
      $500\times$, and $5\times$ against SAC for Inverted Pendulum, Pendulum
      Swing Up, Mountain Car, and Cart Pole, respectively.
    \item Our method requires significantly fewer parameters than state of the
      art model-based methods.  We need at most $n \cdot F$ parameters,
      where $n$ is the dimensionality of the state space and $F$ the number of
      features from which SINDy can choose.
      In comparison, the MBPO network used 613,036 parameters.
    \item Our dynamics models are more interpretable, working from intuitively
      selected or approximated kernel functions, and extracting the governing
      physics dynamics equations and their parameters (coefficients).
\end{enumerate}

A future direction for this work is exploring more complex robotic systems
supported by the Mujoco physical simulation framework.

\section*{Acknowledgements}

The authors would like to thank Lucas N. Alegre for his implementation of the
MBPO algorithm used in this paper.  Rushiv Arora is supported by a Bay State
Fellowship.

{\raggedright
  \bibliography{refs}
  \bibliographystyle{icml2022}
}

\end{document}